\pdfoutput=1

\documentclass[11pt]{article}

\usepackage[preprint]{acl}

\usepackage{times}
\usepackage{latexsym}

\usepackage[T1]{fontenc}

\usepackage[utf8]{inputenc}

\usepackage{microtype}

\usepackage{inconsolata}

\usepackage{graphicx}

\title{Normalized Narrow Jump To Conclusions: Normalized Narrow Shortcuts for Parameter Efficient Early Exit Transformer Prediction}

\author{Amrit Diggavi Seshadri \\
    Sudarshantech Software \\
  \texttt{amrit@sudarshantechsoftware.com}}

\begin{document}
\maketitle
\begin{abstract}
\end{abstract}
With the size and cost of large transformer-based language models growing, recently, there has been interest in shortcut casting of early transformer hidden-representations to final-representations for cheaper model inference. In particular, shortcutting pre-trained transformers with linear transformations over early layers has been shown to improve precision in early inference. However, for large language models, even this becomes computationally expensive. In this work, we propose \textit{Narrow Jump to Conclusions (NJTC)} and \textit{Normalized Narrow Jump to Conclusions (N-NJTC)} - parameter efficient alternatives to standard linear shortcutting that reduces shortcut parameter count by over 97\%. We show that N-NJTC reliably outperforms Identity shortcuts at early stages and offers stable precision from all transformer block levels for GPT-2-XL, Phi3-Mini and Llama2-7B transformer models, demonstrating the viability of more parameter efficient short-cutting approaches.
\section{Introduction}
\label{section1}
\begin{figure}[h]
\centering
\includegraphics[width=\linewidth]{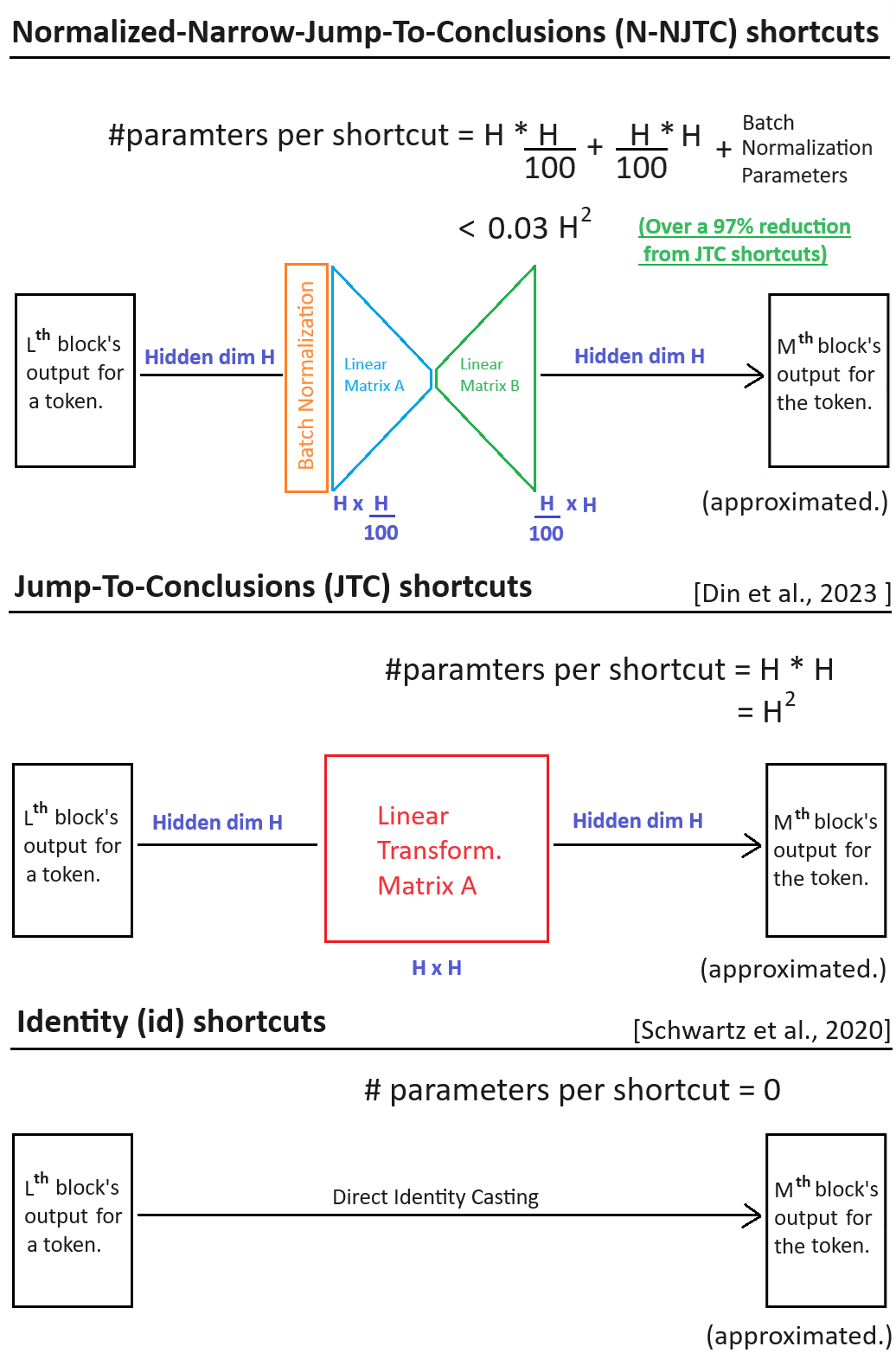}
    \caption{Illustration of our N-NJTC short-cut approache in comparison to previous methods.}
    \label{fig:illustration}
\end{figure}
Transformer based large language models \cite{vaswani2017attention} stack blocks made up of multi-headed self-attention and feed forward layers sequentially. Modern sophisticated language models stack upwards of 30 such blocks. For example, Phi3-Mini \cite{abdin2024phi} stacks 32 transformer blocks, GPT2-XL \cite{radford2019language} stacks 48 blocks, and deeper models like Llama-2 70B \cite{touvron2023llama} and
 GPT-3 \cite{brown2020language} stack as many as 80 and 96 sequential blocks. However while such stacking typically improves model performance, it also increases the computational costs of inference. More GPU memory is required to store the additional stacked transformer blocks and more time is required to forward-pass inputs sequentially. 
\\\\
There have been attempts to reduce the computational costs of such large language models by short-cutting transformers during inference. In short-cutting, one makes intermediate-predictions from an approximation of the final transformer output - that can be cheaply inferred from intermediate transformer-block outputs at each stage in the forward pass. A decision to `early-exit' from the forward pass is made once the confidence of these intermediate-predictions reaches a certain pre-set confidence level $\lambda$. 
\\
Initial methods to short-cutting \cite{schwartz2020right, geva2022transformer} approximated the final transformer output directly by intermediate representations (Identity shortcuts (id) : Figure \ref{fig:illustration}). More recently, \cite{din2023jump} proposed Jump-To-Conclusions (JTC) shortcuts that demonstrate that significant gains in this early-exit inference can be achieved by using a simple linear transformation over token representations to approximate the final transformer output.
\\\\
However, the JTC linear transformation proposed by \cite{din2023jump} adds an additional H x H parameters for each short-cut inference (for a transformer hidden dimension of H). For deep language models with large hidden dimensions, this too becomes very computationally expensive. For example, if we are to shortcut a Phi3-Mini model, we use a transformer hidden dimension of 3072. So each JTC shortcut roughly requires 9.43 Million new parameters. With 32 transformer-blocks, having a shortcut inference option from each block requires that we train and store 9.43 * 31 > 292 Million new parameters. For larger and deeper models like Llama 2 70B, this number grows to over 5 Billion new shortcut parameters.
Clearly, with the size and depth of large transformer models increasing, there is a need to develop more parameter efficient alternatives for short-cutting  than the JTC method.
\\\\
Independent of short-cutting methods, there has also been interest in matrix-decomposition for pre-trained model compression \cite{lan2019albert, noach2020compressing} and success in low-rank fine-tuning of transformer weights \cite{hu2021lora}. LoRA \cite{hu2021lora} in particular has gained traction as a reliable method for parameter efficient fine-tuning - demonstrating that the weight update matrix can be decomposed into low-rank representations to reduce the total number of traninable parameters and reduce costs. However, these prior works focus on increasing efficiency within individual transformer blocks without skipping any block-computations. They do not consider any applications to transformer short-cutting - that skips multiple transformer block computations at a time, and is in itself a method for extreme efficiency.
\\\\
Taking inspiration from \cite{noach2020compressing} and \cite{hu2021lora}, in this work we address the parameter inefficiency of JTC shortcuts \cite{din2023jump}.
\begin{itemize}
    \item We propose Narrow Jump to Conclusions (NJTC) and Normalized Narrow Jump to conclusions (N-NJTC) for shortcutting of transformers - showing that linear shortcuts from early stages can themselves be approximated by low rank representations to achieve over a 97\% parameter reduction from JTC shortcuts.
    
    \item We show that N-NJTC reliably outperforms Identity shortcuts at early stages and offers stable precision from all transformer block levels for GPT-2-XL, Phi3-Mini and Llama2-7B, demonstrating the viability of more parameter efficient short-cutting approaches.
\end{itemize}
\section{Related Work}
As mentioned in Section \ref{section1}, \cite{schwartz2020right} was the first to propose shortcuts for early exit transformer prediction. However they make the assumption that all transformer block outputs operate in the same space and use direct identity shortcuts for prediction. \cite{din2023jump} consider the output of different transformer blocks to operate in different representational spaces and recently demonstrated that linear transformation shortcuts significantly improve the performance of early-exit prediction. However, as discussed, they use full $H \times H$ linear matrices and are not very parameter efficient. Independent of these works \cite{lan2019albert, noach2020compressing} considered matrix-decomposition for pre-trained model compression and  \cite{hu2021lora} demonstrated that training low rank matrix decompositions of a weight update matrix approximate good fine-tuning results for transformers with parameter-efficiency. However as mentioned earlier these methods focus on efficiency within fixed transformer-blocks and do not consider any applications to transformer short-cutting - that skips computations of entire blocks at a time, and is itself a method for extreme efficiency. 
\section{Method}
\subsection{Narrow Jump To Conclusions (NJTC)}
\label{NJTC}
Given a transformer model with hidden dimension $H$, to approximate a short-cut between its block-outputs at any two levels $l$ and $m$, given a set of $N$ input sentences $S$, we forward pass each sentence $s_i \in S$ through the transformer to obtain  intermediate representation pairs $\{(h^l_i, h^m_i)\}_{i=1}^N$  after blocks $l$ and $m$ at randomly selected token positions in each $s_i$. We then fit a simple 2 layer linear neural network made up of matrices $A: H \times \frac{H}{100}$ and $B: \frac{H}{100} \times H$ that takes as input $h^l_i$ and approximates $\hat{h}^m_i$.\\
\begin{equation}
\label{eq-dim}
    \hat{h}^m_i = (h^l_i)AB
\end{equation}
\\
We note in particular that while other model informed-choices for low-rank short-cutting dimensions may be possible, in Eq.\ref{eq-dim} to standardize our approach to diverse transformers with potentially different hidden dimensions, we consider a fixed low-rank reduction to 1\% of the transformer hidden dimension size ($\frac{H}{100}$).
\\\\
We fit the two matrices $A$ and $B$ jointly using gradient descent to minimize the mean squared error loss $L_{lm}$ between the approximated representations $\hat{h}^m_i$ and the transformer block outputs $h^m_i$.\\
\begin{equation}
    L_{lm} = \frac{1}{N}\sum_{i=1}^{N} ||\hat{h}^m_i - h^m_i||^2 
\end{equation}
The hidden representation of each token in a sentence at level $l$ is passed through $A$ and $B$ to obtain approximations of the hidden representations at level $m$.
As a result of this low rank matrix decomposition, each NJTC shortcut uses $2*(H \times \frac{H}{100}) = 0.02H^2$ parameters: Only $2\%$ the number of parameters of a JTC shortcut.
\begin{figure*}
    \centering
    \includegraphics[width=\linewidth]{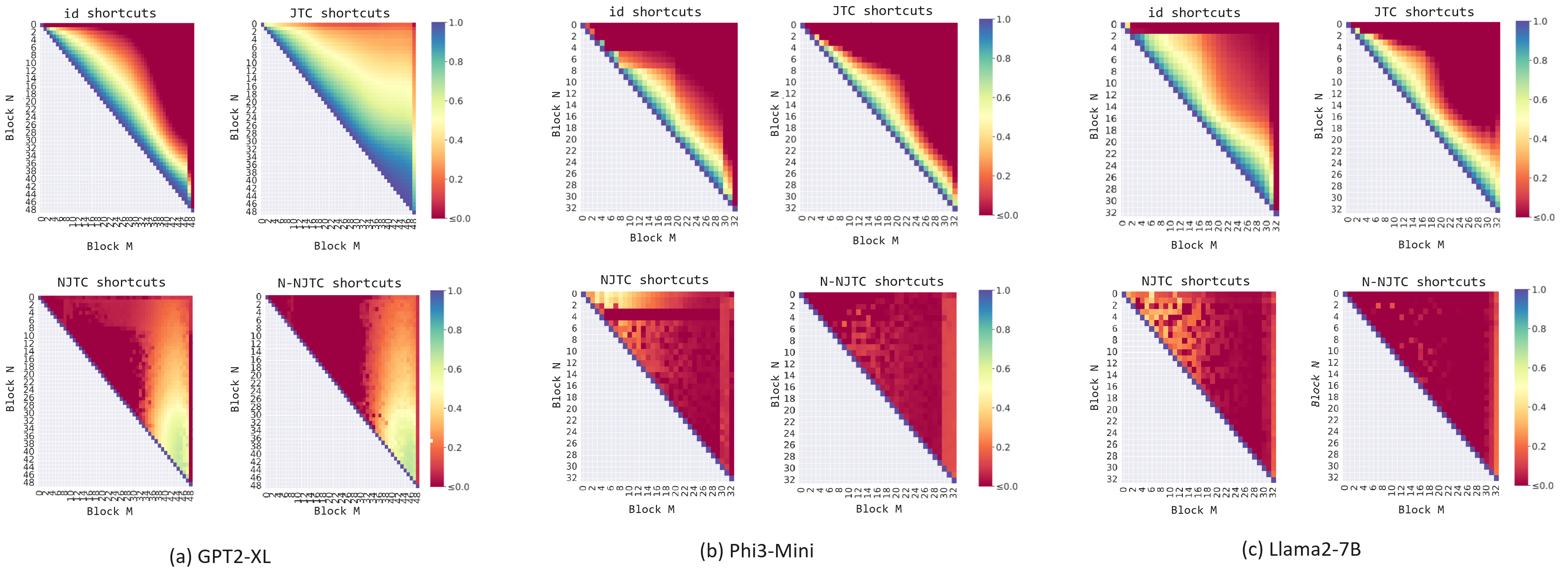}
    \caption{Coordinate-averaged r2-scores ($\uparrow$) between real outputs from transformer Block M and shortcut jump approximations to M from N for different shortcuts types for (a) GPT2-XL, (b) Phi3-Mini and (c) Llama2-7B.}
    \label{combine}
\end{figure*}
\subsection{Normalized Narrow Jump To Conclusions (N-NJTC)}
We note that our NJTC method can be viewed as a special form of a linear denoising auto-encoder - where, the `corrupted input' is a transformer's early block hidden representation $h^l_i$ and the restoration target is a block's output $h^m_i$ further down the forward pass. 
Linear autoencoders usually learn latent dimensions that maximize feature variance and are preceeded by normalization along the batch-dimension to avoid any bias towards naturally high-variance features. Motivated by this comparison, we propose a normalized version of NJTC where we add a batch normalization layer before AB (Fig. \ref{fig:illustration}).
Batch Normalization adds an additional 4H parameters for each shortcut. For  hidden dimension $H>400$, this is less than $0.01H^2$. As all transformer models use a hidden dimension larger than 400, we find that N-NJTC uses less than $3\%$ the number of parameters of a JTC shortcut - offering over a 97\% parameter reduction.
\section{Experiments}
We test our shortcuts on GPT2-XL \cite{radford2019language} which consists of 48 transformer blocks, hidden dimension of 1600, and a total of 1.5 Billion model parameters; on the larger Phi3-Mini \cite{abdin2024phi}  which uses 32 transformer blocks, has hidden dimension 3072, and has a total of 3.8 Billion parameters; and on the even larger Llama2-7B \cite{touvron2023llama} that uses 32 transformer blocks, has hidden dimension 4096 and uses a total  of 7 Billion parameters. The low-rank dimensions $\lfloor\frac{H}{100}\rfloor$ that we use for our NJTC and N-NJTC shortcuts for GPT2-XL, Phi3-Mini and Llama2-7B are $16$ and $30$ and $40$ respectively. 
\\\\
\textbf{Data:} Following the approach taken by \cite{din2023jump}, we sample random sentences from Wikipedia, collecting 9,000 train sentences
and 3000 validation sentences - each of which are highly diverse, written by different authors on varied topics. As explained in Section \ref{NJTC}, each sentence is forward passed through a given transformer model and random token position representations are selected across all hidden representations to train and evaluate shortcuts.
\subsection{Quality of Shortcut Approximations}
We first examine the degree of correlation between true transformer block outputs and their shortcut approximations for each shortcut type. For this purpose, we compute the coordinate averaged r2 scores between true transformer Block M outputs and corresponding shortcut jump approximations made from Block N outputs to Block M. Figure \ref{combine} shows heatmaps of these scores across all transformer block levels for id, JTC, NJTC and N-NJTC shortcuts for GP2-XL, Phi3-Mini and Llama2-7B models respectively.
\\\\
For id and JTC shortcuts, as one would expect, correlation of approximations seems to worsen as jump distance increases. That is, we always achieve better correlated approximations by making a shortcut jump from Block $N$ to Block $(N+1)$ than we could achieve by making a jump to a later Block $(N+2)$. Interestingly, for NJTC and N-NJTC shortcuts, that is not the case. As shown in Figure \ref{combine}, with some exceptions, we typically achieve better correlated approximations by jumping from any intermediate block $N$ directly to the final few blocks output than we could achieve by making a smaller jump from block $N$ to $(N+1)$ at earlier stages. This is an important finding as making jumps to the final block output is all that we really care about for early exit transformer prediction. We are happy to sacrifice intermediate jump quality to improve parameter efficiency of our shortcuts, provided that jumps to the final block outputs are still well correlated with the true final outputs. In this context, we note that N-NJTC shortcuts usually provide better correlated approximation in the final blocks than NJTC shortcuts can.
\subsection{Quality of Shortcut Predictions}
\label{prediction}
\begin{figure}[h]
    \centering
    \includegraphics[width=\linewidth]{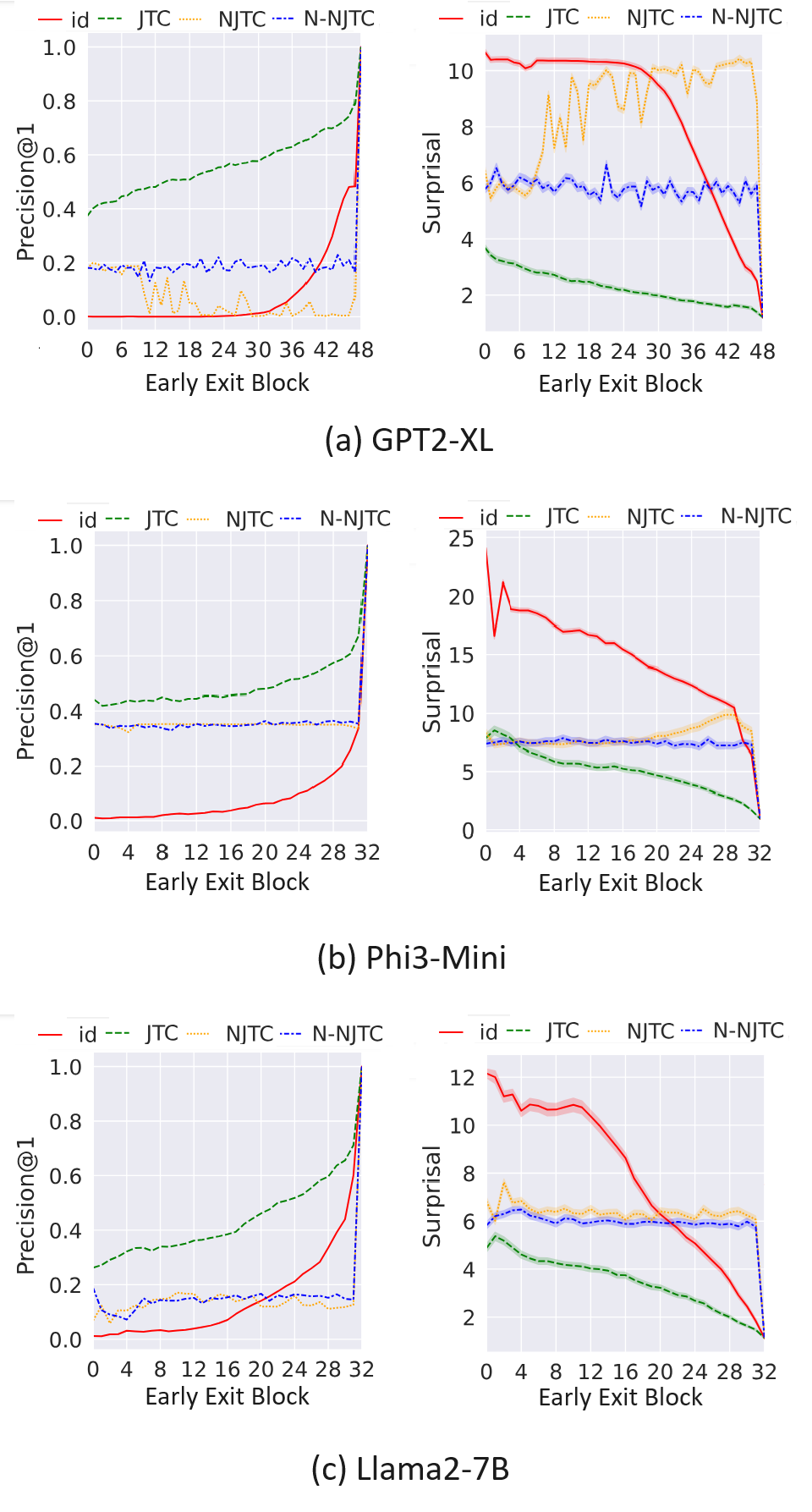}
    \caption{Precision $(\uparrow)$ and Surprisal $(\downarrow)$ achieved by different shortcut types for early exit prediction for (a) GPT-2-XL, (b) Phi3-Mini and (c) Llama2-7B.}
    \label{ps-combine}
\end{figure}

We next consider the quality of shortcut approximations for next token predictions obtained by shortcut jumping to the final transformer block output from each intermediate block. Following the approach taken by \cite{din2023jump}, we compute Precision by assigning a score of 1 if the most-likely token from the shortcut predicted distribution matches the most-likely token from the true final block output distribution and 0 otherwise; and compute Surprisal by measuring the negative log-likelihood of the true block most likely output token according to the shortcut predicted distribution. Figure \ref{ps-combine} shows these Precision and Surprisal scores achieved by id, JTC, NJTC and N-NJTC shortcuts when making an early exit from each transformer block for GPT-2XL, Phi3-Mini and Llama2-7B models respectively. \\As expected, reducing parameter count by $97\%$, NJTC and N-NJTC shortcuts record lower precision and higher surprisal scores than JTC shortcuts for all models. However, their behaviour is still unexpected and interesting.
Our main contribution is the finding that despite the drastic $97\%$ reduction in parameter count from JTC shortcuts, N-NJTC is still able to reliably outperform Identity shortcuts (id) at early transformer model stages. As shown in Figure \ref{ps-combine}, N-NJTC acheives steady and non-fluctuating scores, recording higher precision and lower surprisal than Identity shortcuts (id) upto at least 50$\%$ of a model's total block depth for all three GPT-2XL, Phi3-Mini and Llama2-7B models. This is not a guaranteed finding. With such a large reduction in paramter count, an intutive expectation is that precision and surprisal would record very poor values or fluctuate greatly and collapse quickly. This does infact happen for NJTC in GPT2-XL prediction (Figure \ref{ps-combine}). However N-NJTC solves this problem and remains steady. 
\\\\
To further highlight the surprising nature of trends being observed in a concrete example, consider the Phi3Mini model and a jump from layer 4 to 32. That shortcut jump skips 87.5\% of transformer block computations, i.e > 3 Billion parameter computations. We would expect an id shortcut that uses hidden dimension of size 3072 to perform badly for the missed computation. Intuitively, we would not expect any improvements to result from compressing that early latent vector down from hidden dimension 3072 to 30 and then decompressing it. We would expect a collapse in performance - resulting from a loss of the information that the early transformer output had. However, contrary to that intuition, we find that for our N-NJTC method, not only is there no collapse in performance, but a significant improvement in early transformer block stages - (Figure 3.b). Further, this surprising trend of improvement holds steady up to at least 50\% of model-depth across diverse transformer models – GPT2-XL, Phi3Mini, Llama2-7B that vary in size, vary in structure, vary in creators, and vary in training data. 
The main contribution of our paper is the surprising finding that the drastic parameter efficiency of our N-NJTC method is indeed viable for transformer short-cutting – given that short-cutting is already a method for extreme computational 
efficiency itself.
\section{Practicality and Future Work} 
In terms of immediate practically, we highlight that our N-NJTC method can offer immediate cost savings in settings where one exits the transformer stack early (before executing 50\% of the model's total block depth) and when computation overhead from a full JTC shortcut is substantial, while the performance from Identity shortcuts is unacceptable. In terms of future work on the other hand, we highlight that to our knowledge, we are the first to examine the problem of parameter inefficiency in JTC shortcuts, and the first to consider parameter efficient shortcutting alternatives – we register our surprise that such drastic improvements up to 97\% less costly are indeed viable in early stages, and expect that this observation will spur future interest in building more variants of parameter efficient transformer shortcutting approaches. 
\section{Conclusion}
In this work, we proposed the \textit{Narrow Jump to Conclusions (NJTC)} and \textit{Normalized Narrow Jump to Conclusions (N-NJTC)} methods for parameter efficient shortcutting of transformer models. We showed that linear shortcuts from early stages can themselves be approximated by low rank representations to achieve over a 97$\%$ parameter reduction from JTC shortcuts. We applied our NJTC and N-NJTC methods to GPT-2-XL, Phi3-Mini and Llama2-7B transformer models and showed that N-NJTC reliably outperforms Identity shortcuts at early transformer model stages while also offering stable precision and surprisal from all transformer block levels, demonstrating the viability of more parameter efficient short-cutting methods than JTC.
\section{Limitations}
Notably, as mentioned in Section \ref{prediction}, our NJTC method collapses for GPT2-XL and while N-NJTC solves this problem, NJTC and N-NJTC both achieve worse precision and surprisal scores than JTC shortcuts for all models, and are outperformed by Identity shorotcuts in late-block shortcutting (Figure \ref{ps-combine}). We note however that these limitations are acceptable in exchange for the $97\%$ reduction in parameter count our N-NJTC method offers while outperforming Identity shortcuts at early transformer model stages. We note that shortcutting of transformers in general can cause unexpected model behaviour and caution that any shortcut approximations be tested for safety.  
\bibliography{custom}
\end{document}